\documentclass[11pt,a4paper]{article}
\usepackage[hyperref]{acl2018}
\usepackage{times}
\usepackage{latexsym}
\usepackage{url}

\aclfinalcopy 

\usepackage{amsfonts}
\usepackage{amsmath}
\usepackage{amssymb}
\usepackage{amsthm}
\usepackage{multirow}
\usepackage{algorithm}
\usepackage{algorithmic}
\usepackage{tikz-dependency}
\usepackage{graphicx}
\usepackage{tikz}
\usepackage{pgfplots}

\makeatletter
\newcommand\argmax{\mathop{\operator@font arg~max}}
\newcommand\argmin{\mathop{\operator@font arg~min}}

\newcommand{\figref}[1]{Figure~\ref{#1}}

\makeatother

\graphicspath{{image/}}

\title{HLT@SUDA at SemEval 2019 Task 1: UCCA Graph Parsing as Constituent Tree Parsing}

\author{Wei Jiang, Zhenghua Li\thanks{$~$Corresponding author, \textcolor{darkblue}{hlt.suda.edu.cn/zhenghua}}, Yu Zhang,  Min Zhang \\
School of Computer Science and Technology, Soochow University, China \\
{\tt \{wjiang0501, yzhang25\}@stu.suda.edu.cn, \{zhli13,minzhang\}@suda.edu.cn} 
}

\date{\today}

\begin{document}

\maketitle

\begin{abstract}

This paper describes a simple UCCA semantic graph parsing approach. 
The key idea is to convert a UCCA semantic graph into a constituent tree, in which extra labels are deliberately designed to mark remote edges and discontinuous nodes for future recovery. 
In this way, we can make use of existing syntactic parsing techniques. 
Based on the data statistics, 
we recover discontinuous nodes directly according to the output labels of the constituent parser and 
use a biaffine classification model to recover the more complex remote edges.   
The classification model and the constituent parser are simultaneously trained under the multi-task learning framework. 
We use the multilingual BERT as extra features in the open tracks. Our system ranks the first place in the six English/German closed/open tracks among seven participating systems. 
For the seventh cross-lingual track, where there is little training data for French, we propose a
language embedding approach to utilize English and German training data, and our result ranks the second place. 



\end{abstract}
\section{Introduction}\label{sec:intro}

\begin{figure}[tb]
    \centering
      \scalebox{.9}{
      \begin{tikzpicture}[level distance=11mm, ->,
          every circle node/.append style={inner sep=0.6pt, black, fill=white, draw}]
        \node (ROOT) [circle] {1}
          child {node (2-1) [circle] {2} edge from parent node[xshift=-2pt, yshift=2pt] {\scriptsize H}}
          child {[anchor=north] node[right of=2-1, node distance=2cm] (2-2) [circle] {3}
          {             
            child {node (3-0) [circle] {4}
            {
              child {node(yinhao) {``}}
            } edge from parent[white] node[left] {\scriptsize  } }
            child {node[right of=3-0, node distance=0.9cm] (3-1) [circle] {5}
            {
              child {node(lch) {lch}}
            } edge from parent[white] node[left] {\scriptsize  } }
            child {[sibling distance=1.2cm] node[right of=3-1, node distance=1.4cm] (3-2) [circle] {6}
            {
              child {node(ging) {ging} edge from parent node[left] {\scriptsize  }}
              child {node(umher) {umher} edge from parent node[left] {\scriptsize  }}
            } edge from parent[white] node[left] {\scriptsize A} }
            child {node[right of=3-2, node distance=1.7cm] (3-3) [circle] {7}
            {
              child {node(und) {und} edge from parent node[right] {\scriptsize  }}
            } edge from parent[white] node[left] {\scriptsize A} }
            child {[anchor=north] node[right of=3-3, node distance=1.1cm] (3-4) [circle] {8}
            {
              child {[anchor=north] node (tastete) {tastete} edge from parent node[left] {\scriptsize  }}
            } edge from parent node[left] {\scriptsize  } }
            child {node[right of=3-4, node distance=1cm] (3-5) [circle] {9}
            {
              child {node (juhao) {.} edge from parent node[left] {\scriptsize  }}
            } edge from parent node[left] {\scriptsize  } }
          } edge from parent[white] node {\scriptsize H } }
          ;
          \draw[bend right,->] (2-2.west) to[out=1, in=180] node [xshift=18pt, yshift=10pt] {\scriptsize U} (3-0);
          \draw[bend right,->] (3-0) to[out=1, in=180] node  {\scriptsize } (yinhao);
          \draw[bend right,->] (3-1) to[out=1, in=180] node  {\scriptsize } (lch);
          \draw[bend right,->] (3-2) to[out=1, in=180] node  {\scriptsize  } (ging);
          \draw[bend right,->] (3-2) to[out=1, in=180] node [left] {\scriptsize  } (umher);
          \draw[bend right,->] (3-3) to[out=1, in=180] node  {\scriptsize  } (und);
          \draw[bend right,->] (2-1) to[out=1, in=180] node [xshift=-1pt, yshift=7pt] {\scriptsize A} (3-1);
          \draw[bend right,->] (2-1) to[out=1, in=180] node [xshift=2pt, yshift=8pt] {\scriptsize P} (3-2);
          \draw[bend right,->] (ROOT) to[out=1, in=180] node [xshift=-1pt, yshift=10pt] {\scriptsize L} (3-3);
          \draw[dashed,->] (2-2.south) to node[xshift=3pt, yshift=6pt] {\scriptsize A} (3-1.east);
          \draw[bend right,->] (ROOT) to[out=1, in=180] node[xshift=2pt, yshift=4pt] {\scriptsize H} (2-2);
          \draw[bend right,->] (2-2) to[out=1, in=180] node[xshift=5pt] {\scriptsize P} (3-4);
          \draw[bend right,->] (2-2) to[out=1, in=180] node[xshift=5pt] {\scriptsize U} (3-5);
          \draw[bend right,->] (3-4.south) to[out=1, in=180] node[right] {\scriptsize  } (tastete.north);
          \draw[bend right,->] (3-5.south) to[out=1, in=180] node {\scriptsize  } (juhao.north);
      \end{tikzpicture}}
      \caption{A UCCA graph example from the German data. The English translation is \underline{`` I went around and groped .} We assign a number to each non-terminal node to facilitate illustration.} 
      \label{fig:ucca-example}
    \end{figure}
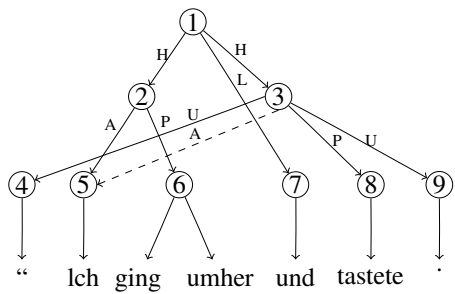

Universal Conceptual Cognitive Annotation (UCCA) is a multi-layer linguistic framework for semantic annotation proposed by  \citet{abend2013universal}.
\figref{fig:ucca-example} shows an example sentence and its UCCA graph. 
Words are represented as terminal nodes. 
Circles denote non-terminal nodes, and the semantic relation between two non-terminal nodes is represented by the label on the edge. 
One node may have multiple parents, among which one is annotated as the primary parent, marked by solid line edges, and others as remote parents, marked by dashed line edges. 
The primary edges form a tree structure, whereas the remote edges  enable reentrancy, forming directed acyclic graphs (DAGs).\footnote{The full UCCA scheme also has implicit and linkage relations, which are overlooked in the community so far.} 
The second feature of UCCA is the existence of nodes with discontinuous leaves, known as discontinuity.  For example, node $3$ in Figure \ref{fig:ucca-example} is discontinuous because some terminal nodes it  spans are not its descendants. 
 
\citet{hershcovich2017a} first propose a transition-based UCCA Parser, which is used as the baseline in the closed tracks of this shared task.  
Based on the recent progress on transition-based parsing techniques, they propose a novel set of transition actions to handle both discontinuous and remote nodes and design useful features based on
bidirectional LSTMs.
\citet{hershcovich2018multitask} then extend their previous approach and propose to utilize the annotated data with other semantic formalisms such as abstract meaning representation (AMR), universal dependencies (UD), and bilexical
Semantic Dependencies (SDP), via multi-task learning, which is used as the baseline in the open tracks.

In this paper, we present a simple UCCA semantic graph parsing approach by treating UCCA semantic graph parsing as constituent parsing. 
We first convert a UCCA semantic graph into a constituent tree by removing discontinuous and remote phenomena. 
Extra labels encodings are deliberately designed to annotate the conversion process and to recover discontinuous and remote structures. 
We heuristically recover discontinuous nodes according to the output labels of the constituent parser, since most discontinuous nodes share the same pattern according to the data statistics. 
As for the more complex remote edges, we use a biaffine classification model for their recovery. 
We directly employ the graph-based constituent parser of \citet{Stern2017Minimal} and jointly train the parser and the biaffine classification model via multi-task learning (MTL).  
For the open tracks, we use the publicly available multilingual BERT as extra features. 
Our system ranks the first place in the six English/German closed/open tracks among seven participating systems. 
For the seventh cross-lingual track, where there is little training data for French, we propose a 
language embedding approach to utilize English and German training data, and our result ranks the second place.


\section{The Main Approach}

Our key idea is to convert UCCA graphs into constituent trees by removing discontinuous and remote edges and using extra labels for their future recovery. 
Our idea is inspired by the pseudo non-projective dependency parsing approach propose by 
\citet{nivre-acl05-pseudo}. 


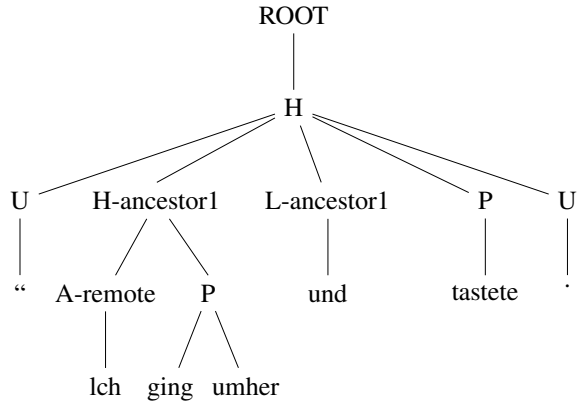
\begin{figure}[tb]
  \centering
    \scalebox{.9}{
    \begin{tikzpicture}[level distance=11mm, -]
      \node (ROOT) {ROOT}
        child {[sibling distance=2cm, anchor=north] node (2-1) {H}
        {
           child {node (4-1) {U}
            { 
                child {node (5-0) {``}}
            }}
          child {[sibling distance=1.5cm, anchor=north] node (3-2) {H-ancestor1}
          {
            child {node (4-2) {A-remote}
            { 
                child {node (5-1) {lch}}
            }}
            child {[sibling distance=1.1cm] node (4-3) {P}
            { 
                child {node (5-2) {ging}}
                child {node (5-3) {umher}}
            }}
          }}
          child {node[right of=3-2, node distance=2.5cm] (3-3) {L-ancestor1}
          {
            child {node (4-4) (und) {und}}
          }}
          child {node[right of=3-3, node distance=2.3cm] (3-4) {P}
          {
            child {node (4-5) {tastete}}
          }}
          child {node (3-5) {U}
          {
            child {node (4-6) {.}}
          }}
        }}
        ;
    \end{tikzpicture}}
    \caption{Constituent tree converted from UCCA gragh.}
    \label{fig:tree}
  \end{figure}
\subsection{Graph-to-Tree Conversion}







Given a UCCA graph as depicted in Figure \ref{fig:ucca-example}, we produce a constituent tree shown in Figure \ref{fig:tree} based on our algorithm described as follows. 

\textbf{1) Removal of remote edges.} 
For nodes that have multiple parent nodes, we remove all remote edges and only keep the primary edge.  
To facilitate future recovery, we concatenate an extra ``remote'' to the label of the primary edge, indicating that the corresponding node has other remote relations. 
We can see that the label of the child node $5$ becomes ``A-remote'' after conversion in Figure \ref{fig:ucca-example} and \ref{fig:tree}. 

\begin{table}
	\scalebox{.9}{
	\begin{tabular}{l|c|c|c|c}
		\hline
		  &  train &  dev  &  total &  percent(\%) \\
		\hline
		ancestor 1 & 1460 & 149 &  1609 & 91.3 \\
        ancestor 2 & 96 & 19 &  115 & 6.5 \\
        ancestor 3 & 21 & 0 &  21 & 1.2 \\
        discontinuous & 16 & 2 & 18 & 1.0 \\
		\hline
	\end{tabular}
	}
	\caption{Distribution of discontinuous structures in the  {English-Wiki} data, which is similar in the German data. 
	}
	\label{table:distributions}
\end{table}



\textbf{2) Handling discontinuous nodes.} 
We call node $3$ in Figure \ref{fig:ucca-example} a discontinuous node because the terminal nodes (also words or leaves) it spans are not continuous (``lch ging umher und'' are not its descendants). 
Since mainstream constituent parsers cannot handle discontinuity, we try to remove discontinuous structures by moving specific edges in the following procedure. 

Given a discontinuous node $A=3$, we first process the leftmost non-descendant node $B=``lch''$.  
We go upwards along the edges until we find a node $C=2$, whose father is either the lowest common ancestor (LCA) of $A=3$ and $B=``lch''$ or another discontinuous node. 
We denote the father of $C=2$ as $D=1$. 

Then we move $C=2$ to be the child of $A=3$, and concatenate the original edge label with an extra string (among ``ancestor 1/2/3/...'' and ``discontinuous'') for future recovery, 
where the number represents the number of edges between the ancestor $D=1$ and $A=3$. 

After reorganizing the graph, we then restart and perform the same operations again until there is no discontinuity. 

Table \ref{table:distributions} shows the statistics of the discontinuous structures in the {English-Wiki} data. 
We can see that $D$ is mostly likely the LCA of $A$ and $B$, and there is only one edge between $D$ and $A$ in more than $90\%$ cases. 

Considering the skewed distribution, we only keep ``ancestor 1'' after graph-to-tree conversion, and treat others as continuous structures for simplicity. 



\textbf{3) Pushing labels from edges into nodes.} 
Since the labels are usually annotated in the nodes instead of edges in constituent trees, we push all labels from edges to the child nodes. 
We label the top node as ``ROOT''.

\tikzset{
wordnode/.style = {circle,fill=black,inner sep=0,minimum size=4pt},
arc/.style = {draw,  -latex'},
decision/.style = {diamond, draw, fill=blue!20,
  text width=4.5em, text badly centered, node distance=3cm, inner sep=0pt},
block/.style = {rectangle, draw, fill=blue!20,
  text width=3em, text centered, rounded corners, minimum height=4em},
textdata/.style = {rectangle,
  text width=3em, text centered, minimum height=4em},
procedure/.style = {rectangle, draw, fill=blue!20,
  text width=3.5em, text centered,  minimum height=3em,  minimum width=3em},
line/.style = {draw,very thick,  -latex'},
line1/.style = {draw, thick, dashed},
compose/.style = {rectangle, draw, fill=none, rounded corners, blue, thick},
cloud/.style = {draw, ellipse,fill=red!20, node distance=3cm,
  minimum height=2em},
subroutine/.style = {draw,rectangle split,
  rectangle split parts=3,minimum height=1cm,
  rectangle split part fill={red!50, green!50, blue!20, yellow!50}},
connector/.style = {draw,circle,node distance=3cm,fill=yellow!20},
data/.style = {draw, trapezium, text width=3em, align=center, fill=olive!20, trapezium left angle=75pt, trapezium right angle=105, trapezium stretches = true, minimum height=3em, minimum width = 3em},
mytrap/.style={trapezium, trapezium angle=67.5, draw, inner ysep=5pt, outer sep=0pt,  minimum height=1.81mm, minimum width=0pt}
}

\definecolor{a1c}{rgb}{0.00,0.00,1}
\definecolor{a2c}{rgb}{0.30,0.00,0.70}
\definecolor{a3c}{rgb}{0.00,0.50,0.50}
\definecolor{a4c}{rgb}{0.70,0.00,0.70}
\definecolor{a5c}{rgb}{1.00,0.00,0.00}
\definecolor{a6c}{rgb}{0.00,1.00,0.00}

\definecolor{redi}{RGB}{255,38,0}
\definecolor{redii}{RGB}{200,50,0}
\definecolor{yellowi}{RGB}{255,251,0}
\definecolor{bluei}{RGB}{0,150,255}
\definecolor{blueii}{RGB}{135,247,210}
\definecolor{blueiii}{RGB}{91,205,250}
\definecolor{blueiv}{RGB}{115,244,253}
\definecolor{bluev}{RGB}{1,58,215}
\definecolor{orangei}{RGB}{240,143,50}
\definecolor{yellowii}{RGB}{222,247,100}
\definecolor{greeni}{RGB}{166,247,166}

\depstyle{stanford1}{edge style = {thick, red}, edge height = 0.3cm, edge horizontal padding=9.0pt, label style={thick, draw=white, text=red, fill=white, rotate = 0}}
\depstyle{stanford2}{edge style = {thick, red}, edge height = 0.6cm, edge horizontal padding=7.2pt, label style={thick, draw=white, text=red, fill=white, rotate = 0}}
\depstyle{stanford3}{edge style = {thick, red}, edge height = 0.9cm, edge horizontal padding=5.4pt, label style={thick, draw=white, text=red, fill=white, rotate = 0}}
\depstyle{stanford4}{edge style = {thick, red}, edge height = 1.6cm, edge horizontal padding=3.6pt, label style={thick, draw=white, text=red, fill=white, rotate = 0}}
\depstyle{stanford5}{edge style = {thick, red}, edge height = 1.9cm, edge horizontal padding=1.8pt, label style={thick, draw=white, text=red, fill=white, rotate = 0}}
\depstyle{stanford6}{edge style = {thick, red}, edge height = 2.5cm, edge horizontal padding=0.0pt, label style={thick, draw=white, text=red, fill=white, rotate = 0}}
\depstyle{stanford7}{edge style = {thick, red}, edge height = 1.5cm, edge horizontal padding=0.0pt, label style={thick, draw=white, text=red, fill=white, rotate = 0}}

\begin{figure}[tb]
\begin{center}
\begin{small}
\begin{tikzpicture}[node distance = 0.1cm, auto]
\definecolor{mycolor}{RGB}{228, 198, 208}
\definecolor{mycolor2}{RGB}{161, 175, 201}
\definecolor{mycolor3}{RGB}{135, 186, 191}

\node [inner sep=0pt] (the_orig) {};


\node [inner sep=0pt, above of=the_orig, node distance = 0.5em] (xi) {\bf $\mathbf{x}_i 
$};

\node [inner sep=0pt, left of=xi, node distance = 7em] (x0) {\bf ...};

\node [inner sep=0pt, right of=xi, node distance = 7em] (x1) {\bf ...};



\node [rectangle, above of =xi, node distance = 4em, rounded corners, draw, fill=red!20, text centered, minimum width=8em, minimum height=1.5em] (lstm) {Shared BiLSTMs};

\node [rectangle, above left = 3em and -2.5em of lstm, rounded corners, draw, fill=blue!40, text centered, minimum width=8em, minimum height=1.5em] (mlp_src) {MLPs and Biaffines};

\node [rectangle, above right = 3em and -2.5em of lstm, rounded corners, draw, fill=blue!40, text centered, minimum width=8em, minimum height=1.5em] (mlp_tgt) {MLPs};

\node [rectangle, rounded corners, draw, fill=mycolor2, text centered, minimum width=3.5em, minimum height=1.5em, above of = mlp_src, node distance = 4em] (biaffine_src) {Remote recovery};

\node [rectangle, rounded corners, draw, fill=mycolor2, text centered, minimum width=3.5em, minimum height=1.5em, above of = mlp_tgt, node distance = 4em] (biaffine_tgt) {Constituent Parsing};

\node [inner sep=0pt, above of = xi, node distance = 0.5em] (above_xi){};

\path [draw, thick, ->, >=stealth'] (above_xi) to [out=90,in=-90]  (lstm);

\path [draw, thick, ->, >=stealth'] (lstm) to [out=90,in=-90]  (mlp_src);

\path [draw, thick, ->, >=stealth'] (lstm) to [out=90,in=-90]  (mlp_tgt);

\path [draw, thick, ->, >=stealth'] (mlp_src) to [out=90,in=-90]  (biaffine_src);

\path [draw, thick, ->, >=stealth'] (mlp_tgt) to [out=90,in=-90]  (biaffine_tgt);

\end{tikzpicture}
\caption{The framework of MTL.}\label{fig:mtl}
\end{small}
\end{center}
\end{figure}
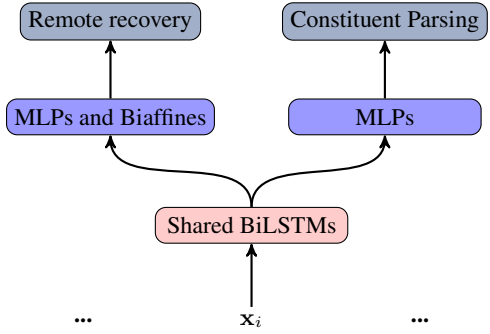
        


\subsection{Constituent Parsing}





We directly adopt the minimal span-based parser of \citet{Stern2017Minimal}. 
Given an input sentence $\mathbf{s}=w_1...w_n$, each word $w_i$ is mapped into a dense vector $\mathbf{x}_i$ via lookup operations. 
$$\mathbf{x}_i = \mathbf{e}_{w_i} \oplus \mathbf{e}_{t_i} \oplus ...$$
where $\mathbf{e}_{w_i}$ is the word embedding and $\mathbf{e}_{t_i}$ is the part-of-speech tag embedding. To make use of other auto-generated linguistic features, provided with the datasets, 
we also include the embeddings of the named entity tags and the dependency labels, but find limited performance gains. 

Then, the parser employs two cascaded bidirectional LSTM layers as the encoder, and use the top-layer outputs as the word representations.  

Afterwards, the parser represents each span $w_i...w_j$ as  
$$\mathbf{r}_{i, j} = (\mathbf{f}_j - \mathbf{f}_i) \oplus (\mathbf{b}_i - \mathbf{b}_j)$$
where $\mathbf{f}_i$ and $\mathbf{b}_i$ are the output vectors of the top-layer forward and backward LSTMs.

The span representations are then fed into MLPs to compute the scores of span splitting and labeling.
For inference, the parser performs greedy top-down searching to build a parse tree. 
\subsection{Remote Edge Recovery}


We borrow the idea of the state-of-the-art biaffine dependency parsing \cite{Timothy-d17-biaffine} and build our remote edge recovery model.
The model shares the same inputs and  LSTM encoder as the constituent parser under the MTL framework \cite{collobert-08-multi-task}. 
For each remote node, marked by ``-remote'' in the constituent tree,  
we consider all other non-terminal nodes as its candidate remote parents. 
Given a remote node $A$ and another non-terminal node $B$, we first represent them as the span representations. $\mathbf{r}_{i,j}$ and $\mathbf{r}_{i',j'}$,
where $i, i', j, j'$ are the start and end word indices governed by the two nodes. 
Please kindly note that $B$ may be a discontinuous node. 

Following \citet{Timothy-d17-biaffine}, we apply two separate MLPs to the remote and candidate parent nodes respectively, producing $\mathbf{r}^{\texttt {child}}_{i,j}$ and 
$\mathbf{r}^{\texttt{parent}}_{i',j'}$. 

Finally, we compute a labeling score vector via a biaffine operation. 
\begin{equation} \label{eq:biaffine}
 \mathbf{s}(A \leftarrow B) =  \left[
 \begin{array}{c}
    \mathbf{r}^{\texttt{child}}_{i, j}    \\
      1 
 \end{array} 
 \right]^\mathrm{T} 
 \mathbf{W}  \mathbf{r}_{i',j'}^{\texttt{parent}} 
\end{equation} 
where the dimension of the labeling score vector is the number of the label set, including a ``NOT-PARENT'' label. 

\textbf{Training loss.} We accumulate the standard cross-entropy losses of all remote and non-terminal node pairs. The parsing loss and the remote edge classification loss are added in the MTL framework. 


 
\subsection{Use of BERT}

For the open tracks, we use the contextualized word representations produced by BERT \cite{devlin2018bert} as extra input features.\footnote{We use the multilingual cased BERT from \url{https://github.com/google-research/bert}.} 
Following previous works, we use the weighted summation of the last four transformer layers and then multiply a task-specific weight parameter following \cite{Peters2018}. 

\section{Cross-lingual Parsing}
Because of little training data for French, we borrow the treebank embedding approach of \citet{Stymne-p18-treebank-emb} for exploiting multiple heterogeneous treebanks for the same language, and propose a language embedding approach to utilize English and German training data. 
The training datasets of the three languages are merged to train a single UCCA parsing model. 
The only modification is to 
concatenate each word position with an extra language embedding (of dimension $50$), i.e. $\mathbf{x}_i \oplus \mathbf{e}_{lang=en/de/fr}$ to indicate which language this training sentence comes from. 
In this way, we expect the model can fully utilize all training data since most parameters are shared except the three language embedding vectors, and learn the language differences as well.
\section{Experiments}

Except BERT, all the data we use, including the linguistic features and word embeddings, are provided by the shared task organizer \citep{daniel-semeval19-ucca}. We also adopt the averaged F1 score as the main evaluation metrics returned by the official evaluation scripts \citep{daniel-semeval19-ucca}.



We train each model for at most 100 iterations, and early stop training if the peak performance does not increase in 10 consecutive iterations. 

\setlength{\tabcolsep}{2pt}
\begin{table}[tb]
\begin{center}
\begin{small}
 \begin{tabular}{c|*{3}{c}}
  \hline
  \multirow{2}{*}{Methods} & \multicolumn{3}{c}{{F1 score}} \\
   & \small Primary & \small Remote & \small Avg \\
  \hline
    \multicolumn{4}{c}{\footnotesize Single-language models on English} \\
    \hline
  random emb 
  & 0.778 & 0.542 & 0.774 \\
  pretrained emb (no finetune) 
  & 0.790 & 0.494 & 0.785 \\
  pretrained emb 
  & 0.794 & 0.535 & \textbf{0.789} \\
  bert 
  & 0.821 & 0.593 & 0.817 \\
  pretrained emb $\oplus$ bert
  & 0.825 & 0.603 & \underline{\textbf{0.821}} \\
  official baseline (closed)	
  & 0.745 & 0.534 & 0.741 \\
  official baseline (open)
  & 0.753 & 0.514 & 0.748 \\
  \hline  \hline
\multicolumn{4}{c}{\footnotesize Single-language models on German} \\
  \hline
  random emb 
  & 0.817 & 0.549 & 0.811 \\
 pretrained emb (no finetune) 
  & 0.829 & 0.544 & 0.823 \\
  pretrained emb 
  & 0.831 & 0.536 & \textbf{0.825} \\
  bert
  & 0.842 & 0.610 & 0.837 \\
  pretrained emb $\oplus$ bert
  & 0.849 & 0.628 & \underline{\textbf{0.844}} \\
   official baseline (closed)
  & 0.737 & 0.46 & 0.731 \\
   official baseline (open)
  & 0.797 & 0.587 & 0.792 \\
  \hline \hline
  \multicolumn{4}{c}{\footnotesize Multilingual models on French} \\
  \hline
   random emb 
  & 0.688 & 0.343 & 0.681 \\
pretrained emb 
  & 0.673 & 0.174 & 0.665 \\
   bert
  & 0.796 & 0.524 & \underline{\textbf{0.789}} \\
   official baseline (open)
  & 0.523 & 0.016 & 0.514 \\
  \hline
 \end{tabular}
\end{small}
\caption{Results on the dev data.}
\label{table:results}
 
\end{center}
\end{table}

Table~\ref{table:results} shows the results on the dev data. 
We have experimented with different settings to gain insights on the contributions of different components. 
For the single-language models, it is clear that using pre-trained word embeddings outperforms using randomly initialized word embeddings by more than 1\% F1 score on both English and German. 
Finetuning the pre-trained word embeddings leads to consistent yet slight performance improvement. 
In the open tracks, replacing word embedding with the BERT representation is also useful on English (2.8\% increase) and German (1.2\% increase). 
Concatenating pre-trained word embeddings with BERT outputs leads is also beneficial. 

For the multilingual models, using randomly initialized word embeddings is better than pre-trained word embeddings, which is contradictory to the single-language results. 
We suspect this is due to that the pre-trained word embeddings are independently trained for different languages and would lie in different semantic spaces without proper aligning. 
Using the BERT outputs is tremendously helpful, boosting the F1 score by more than 10\%. 
We do not report the results on English and German for brevity since little improvement is observed for them. 



\section{Final Results}

Table~\ref{table:test-results} lists our final results on the test data. 
Our system ranks the first place in six tracks (English/German closed/open) and the second place in the French open track. Note that we submitted a wrong result for the French open track during the evaluation phase by setting the wrong index of language, which leads to about 2\% drop of averaged F1 score (0.752). Please refer to  \cite{daniel-semeval19-ucca} for the complete results and comparisons.

\setlength{\tabcolsep}{2pt}
\begin{table}[tb]
\begin{center}
\begin{small}
 \begin{tabular}{c|*{3}{c}}
  \hline
  \multirow{2}{*}{Tracks} & \multicolumn{3}{c}{{F1 score}} \\
   & \small Primary & \small Remote & \small Avg \\
  \hline
  English-Wiki\_closed 
  & 0.779 & 0.522 & 0.774 \\
  English-Wiki\_open 
  & 0.810 & 0.588 & 0.805 \\
  English-20K\_closed 
  & 0.736 & 0.312 & 0.727 \\
  English-20K\_open
  & 0.777 & 0.392 & 0.767 \\
  German-20K\_closed
  & 0.838 & 0.592 & 0.832 \\
  German-20K\_open	
  & 0.854 & 0.641 & 0.849 \\
  French-20K\_open
  & 0.779 & 0.438 & 0.771 \\
  \hline
 \end{tabular}
\end{small}
\caption{Final results on the test data in each track. Please refer to the official  webpage  for more detailed results
due to the limited space}
\label{table:test-results}
 
\end{center}
\end{table}
\section{Conclusions}
In this paper, we describe our system submitted to SemEval 2019 Task 1. 
We design a simple UCCA semantic graph parsing approach by making full use of the recent advance in syntactic parsing community. The key idea is to convert UCCA graphs into constituent trees. 
The graph recovery problem is modeled as another classification task under the MTL framework. 
For the cross-lingual parsing track, we design a language embedding approach to utilize the training data of resource-rich languages.



\section*{Acknowledgements}

The authors would like to thank the anonymous
reviewers for the helpful comments. We also thank
Chen Gong for her help 
on speeding up the minimal span parser. 
This work was supported by National Natural Science Foundation of China (Grant
No. 61525205, 61876116).

\bibliographystyle{acl_natbib}
\bibliography{reference}

\begin{thebibliography}{11}
\expandafter\ifx\csname natexlab\endcsname\relax\def\natexlab#1{#1}\fi

\bibitem[{Abend and Rappoport(2013)}]{abend2013universal}
Omri Abend and Ari Rappoport. 2013.
\newblock {U}niversal {C}onceptual {C}ognitive {A}nnotation ({UCCA}).
\newblock In \emph{Proc. of ACL}, pages 228--238.

\bibitem[{Collobert and Weston(2008)}]{collobert-08-multi-task}
Ronan Collobert and Jason Weston. 2008.
\newblock A unified architecture for natural language processing: Deep neural
  networks with multitask learning.
\newblock In \emph{Proc. of ICML}.

\bibitem[{Devlin et~al.(2018)Devlin, Chang, Lee, and
  Toutanova}]{devlin2018bert}
Jacob Devlin, Ming-Wei Chang, Kenton Lee, and Kristina Toutanova. 2018.
\newblock Bert: Pre-training of deep bidirectional transformers for language
  understanding.
\newblock \emph{arXiv:1810.04805}.

\bibitem[{Dozat and Manning(2017)}]{Timothy-d17-biaffine}
Timothy Dozat and Christopher~D. Manning. 2017.
\newblock Deep biaffine attention for neural dependency parsing.
\newblock In \emph{Proceedings of ICLR}.

\bibitem[{Hershcovich et~al.(2017)Hershcovich, Abend, and
  Rappoport}]{hershcovich2017a}
Daniel Hershcovich, Omri Abend, and Ari Rappoport. 2017.
\newblock A transition-based directed acyclic graph parser for ucca.
\newblock In \emph{Proc. of ACL}, pages 1127--1138.

\bibitem[{Hershcovich et~al.(2018)Hershcovich, Abend, and
  Rappoport}]{hershcovich2018multitask}
Daniel Hershcovich, Omri Abend, and Ari Rappoport. 2018.
\newblock Multitask parsing across semantic representations.
\newblock In \emph{Proc. of ACL}, pages 373--385.

\bibitem[{Hershcovich et~al.(2019)Hershcovich, Aizenbud, Choshen, Sulem,
  Rappoport, and Abend}]{daniel-semeval19-ucca}
Daniel Hershcovich, Zohar Aizenbud, Leshem Choshen, Elior Sulem, Ari Rappoport,
  and Omri Abend. 2019.
\newblock Semeval 2019 task 1: Cross-lingual semantic parsing with ucca.
\newblock \emph{arXiv:1903.02953}.

\bibitem[{Nivre and Nilsson(2005)}]{nivre-acl05-pseudo}
Joakim Nivre and Jens Nilsson. 2005.
\newblock Pseudo-projective dependency parsing.
\newblock In \emph{Proc. of ACL}, pages 99--106.

\bibitem[{Peters et~al.(2018)Peters, Neumann, Iyyer, Gardner, Clark, Lee, and
  Zettlemoyer}]{Peters2018}
Matthew~E. Peters, Mark Neumann, Mohit Iyyer, Matt Gardner, Christopher Clark,
  Kenton Lee, and Luke Zettlemoyer. 2018.
\newblock Deep contextualized word representations.
\newblock In \emph{Proc. of NAACL}.

\bibitem[{Stern et~al.(2017)Stern, Andreas, and Klein}]{Stern2017Minimal}
Mitchell Stern, Jacob Andreas, and Dan Klein. 2017.
\newblock A minimal span-based neural constituency parser.
\newblock In \emph{Proc. of ACL}, pages 818--827.

\bibitem[{Stymne et~al.(2018)Stymne, de~Lhoneux, Smith, and
  Nivre}]{Stymne-p18-treebank-emb}
Sara Stymne, Miryam de~Lhoneux, Aaron Smith, and Joakim Nivre. 2018.
\newblock Parser training with heterogeneous treebanks.
\newblock In \emph{Proc. of ACL}, pages 619--625.

\end{thebibliography}
\end{document}